\documentclass[sigconf]{acmart}
\usepackage{booktabs} 
\usepackage[utf8x]{inputenc} 
\usepackage[T1]{fontenc}    
\usepackage{makecell}
\usepackage{enumitem}
\usepackage{graphicx}
\usepackage{caption}
\usepackage{subcaption}
\usepackage{hyperref}       
\usepackage{url}            
\usepackage{booktabs}       
\usepackage{amsfonts}       
\usepackage{nicefrac}       
\usepackage{microtype}      
\usepackage{enumitem}
\usepackage{multirow}
\usepackage{float}

\acmDOI{10.475/123_4}

\acmISBN{123-4567-24-567/08/06}

\acmConference[Arxiv]{}{April 2018}{}
\acmYear{2018}
\copyrightyear{2018}

\acmPrice{15.00}

\begin{document}
\title{eCommerceGAN : A Generative Adversarial Network for E-commerce}

\author{Ashutosh Kumar}
\authornote{Work done as an intern at Amazon India Machine Learning, Bangalore}
\affiliation{%
  \institution{Indian Institute of Science}
  \city{Bangalore}
}
\email{ashutosh@iisc.ac.in}

\author{Arijit Biswas}
\affiliation{%
  \institution{Amazon India Machine Learning}
  \city{Bangalore}
}
\email{barijit@amazon.com}

\author{Subhajit Sanyal}
\affiliation{%
  \institution{Amazon India Machine Learning}
  \city{Bangalore}
}
\email{subhajs@amazon.com}


\begin{abstract}
E-commerce companies such as Amazon, Alibaba and Flipkart process billions of orders every year. However, these orders represent only a small fraction of all plausible orders. Exploring the space of all plausible orders could help us better understand the relationships between the various entities in an e-commerce ecosystem, namely the customers and the products they purchase. In this paper, we propose a Generative Adversarial Network (GAN) for orders made in e-commerce websites. Once trained, the generator in the GAN could generate any number of plausible orders. Our contributions include: (a) creating a dense and low-dimensional representation of e-commerce orders, (b) train an ecommerceGAN (ecGAN) with real orders to show the feasibility of the proposed paradigm, and (c) train an ecommerce-conditional-GAN (ec\textsuperscript{2}GAN) to generate the plausible orders involving a particular product. We propose several qualitative methods to evaluate ecGAN and demonstrate its effectiveness. The ec\textsuperscript{2}GAN is used for various kinds of characterization of possible orders involving a product that has just been introduced into the e-commerce system. The proposed approach ec\textsuperscript{2}GAN performs significantly better than the baseline in most of the scenarios. 
\end{abstract}

\keywords{E-commerce, Generative Adversarial Networks, Deep Learning, Order Embedding, Product Recommendation}


\maketitle

\section{Introduction}

Major e-commerce companies such as Amazon, Alibaba, Flipkart and eBay have billions of products in their inventories. However, the space of products is of much lower dimension as there is an inherent underlying structure imposed by product categories, sub-categories, price ranges, brands, manufacturers, etc. Similarly on the customer side, though there are several hundreds of millions of customers, they reside in a lower dimensional space as customers can be grouped together based on similarity of purchase behaviors, price/brand sensitivity, ethnicity, etc. Thus the space of e-commerce orders is an interaction of samples from these two spaces where certain interactions are plausible while others are unlikely to happen ever. For example, it is unlikely for a user,  who ``belongs'' to a group of users who are jazz aficionados, to purchase a heavy metal album. On the other hand, if a new tool-kit is introduced in the market, the users who are likely to be the first ones to buy it are the ones who belong to the ``Do it Yourself (DIY)'' group of users. However, the real orders, i.e., the orders which have been placed in an e-commerce website represent only a tiny fraction of all plausible orders. Exploring the space of all plausible orders could provide important insights into product demands, customer preferences, price estimation, seasonal variations etc., which, if taken into consideration, could directly or indirectly impact revenue and customer satisfaction. 

In this paper, we propose an approach to learn the distribution of plausible orders using a Generative Adversarial Network~\cite{NIPS2014_5423}. Generative adversarial network (GAN) is a variation of deep neural network that learns a generative model given a set of real data points. The network constitutes of two components: a generator and a discriminator. The generator takes a random noise vector as input and tries to generate samples which are similar to the real data points. Whereas, the discriminator tries to differentiate between the generated fake data points and the real data points. The generator and the discriminator compete with each other and while doing that the generator ends up learning the real data distribution. In our case, the generator learns to generate many novel yet plausible orders which were never ordered in the e-commerce website. 

This paper makes three major contributions:
\begin{itemize}[noitemsep, leftmargin=*]
\item {\bf Order Representation:} It is important to learn dense, low dimensional and semantically meaningful representations of orders for using them in a generative network such as GAN. We represent each order as a tuple of \{customer, product, price, date\}. Each component in the tuple is represented as a dense and low-dimensional vector and they are concatenated to represent an order. This is described in details in Section \ref{order_representation}.
\item {\bf ecommerceGAN (ecGAN):} In this version, we use a Wasserstein GAN~\cite{arjovsky2017towards}  (Section \ref{WGAN}) to train a generator which can generate plausible e-commerce orders. We train the ecGAN to explore the overall viability of the proposed approach. More details on this is provided in Section \ref{ecgan}.
\item {\bf ecommerce-conditional-GAN (ec\textsuperscript{2}GAN):} In this version, we propose a conditional GAN to generate orders which are conditioned on a particular product, i.e., its representation. In the e-commerce space, it is desirable to understand the characteristics of future orders involving a particular product, especially when a new product is launched. As an example, being able to estimate the demographics of the potential customer-base for a new product can help companies make more informed decisions about their marketing efforts. If we can predict the gender, age, tenure, location, and purchase volumes of possible customers accurately, they can be targeted with personalized deals and recommendations. In a similar vein, better estimates of demand and sales can bolster inventory and supply chain decisions. We propose a variation of Wasserstein GAN, called as ecommerce-conditional-GAN or ec\textsuperscript{2}GAN, where the generator takes a product representation as input (along with random noise vectors) and generates many plausible orders involving this particular product. We add an additional component to the generator loss, which essentially reconstructs the input product representation. More details on this is provided in Section \ref{ec2gan}. 
\end{itemize}

Evaluation of GANs is difficult~\cite{theis2015note, wu2016quantitative, quantitative_eval_gan}, especially when applied to non-visual domains. We propose to use three different methods to qualitatively evaluate the learned generator in ecGAN: t-SNE~\cite{van2008visualizing}, data distribution in the leaves of random forests, and feature correlation. We demonstrate that it is possible to effectively learn the distribution of e-commerce orders. We evaluate ec\textsuperscript{2}GAN quantitatively and demonstrate that this could be used to effectively characterize customers, prices and seasonal demands, when a new product is launched in an e-commerce website. We compare the proposed ec\textsuperscript{2}GAN with a Conditional Variational Autoencoder (C-VAE)~\cite{C-VAE} based order generation approach, and show that ec\textsuperscript{2}GAN outperforms the baseline in seven out of the nine applied use-cases (maximum absolute improvement is $\sim22\%$).    

\section{Related Work}
Generative Adversarial Networks~\cite{NIPS2014_5423} have gained significant popularity in the machine learning community recently as it provides a novel and easy way to learn generative models. It has shown encouraging results in application domains such as computer vision and natural language processing. In \cite{radford2015unsupervised}, the authors propose deep convolutional generative adversarial networks (DCGAN), which have certain architectural constraints and are effective for unsupervised learning. They show that the deep convolutional adversarial pair learns a hierarchy of representations from object parts to scenes in both the generator and the discriminator. The authors in \cite{denton2015deep} use a cascade of convolutional networks within a Laplacian pyramid framework to generate images in a coarse-to-fine fashion. In \cite{zhu2017unpaired}, the authors propose an adversarial network which can learn image-to-image translation without using any paired data. Professor forcing~\cite{lamb2016professor} applied GANs along with RNNs as generative units to model sequential data which are prevalent in the speech and NLP domain. Recently GANs have also been used in the context of topic modeling~\cite{glover2016modeling}. The proposed model is based on the Energy-Based GANs~\cite{DBLP:journals/corr/ZhaoML16}, but uses a Denoising Autoencoder as the discriminator network.

There are several prior works on improving the training, interpretability of noise, and optimization in GANs. InfoGAN~\cite{chen2016infogan} is an information-theoretic extension to the GANs that is able to learn disentangled representations in an unsupervised manner. In \cite{salimans2016improved}, the authors present various architectural features and training procedures which can be applied to improve the training of GANs. GANs are known to be unstable, difficult to train and plagued with the problem of mode-collapse\footnote{In mode collapse, several noise vectors point to the same output vector, which significantly decreases the diversity of samples generated.}. Recently proposed Wasserstein GAN (WGAN)~\cite{DBLP:conf/icml/ArjovskyCB17} improves stability issues and mitigates the problem of mode-collapse to some extent. To capture the distance between the actual and generated distribution, it uses Earth-Movers distance as a metric as opposed to the conventional Jensen-Shannon divergence. The authors use ideas from reinforcement learning, measure theory and adopt a novel discriminator loss, which is referred to as \emph{critic loss} in the paper. The results are comparable to the state-of-the-art, with fewer hyper-parameters and stronger theoretical guarantees. A follow up work~\cite{gulrajani2017improved} which comes as an improvement over Wasserstein GANs alleviates the problem of hyper-parameter tuning even further, keeping the stability intact. Gradient penalty is used along with WGAN to remove instability in some settings, which are typically introduced due to critic weight clipping in WGAN.

To the best of authors' knowledge there is no prior work that applied GANs in the e-commerce domain. This is the first work that learns a generator for plausible e-commerce orders and demonstrates applications on product-conditioned order characterization. We also propose several novel quantitative and qualitative ways to evaluate GANs, which would be applicable to many other domains as well.

\section{Proposed Approach}

In this section, we describe the details of the proposed approaches: ecGAN and ec\textsuperscript{2}GAN. In Section \ref{GAN} and \ref{WGAN}, we describe the traditional GAN and WGAN respectively. In Section \ref{order_representation}, we describe the proposed method, which represents e-commerce orders using dense and low-dimensional vectors. In Section \ref{ecgan} and \ref{ec2gan}, the details of the proposed GAN variations are provided.

\subsection{Generative Adversarial Network}
\label{GAN}

Generative adversarial networks (GAN)~\cite{NIPS2014_5423} are a variation of deep neural network that are believed to learn a generative model given a bunch of real data points. The network comprises of two components: a generator and a discriminator. The generator takes a random noise vector as input and tries to generate samples which are similar to the real data points. Whereas, the discriminator tries to differentiate between the generated data points and the real data points. The generator and the discriminator compete with each other and in that process the generator learns the real data distribution.

Let the real distribution of the data be $p_{\text{Data}}(x)$. Let the distribution of the noise parameter (input to the generator) be $p_Z(z)$. We use the standard normal distribution for the noise parameters. The discriminator function is defined as $D: X \to [0,1]$, where $X$ is the input to the discriminator. The Generator function is defined as, $G:Z \to X'$, where $Z$ is the input noise and $X'$ is the output of the generator, the dimension of which is equal to the dimension of feature vector of the real data points.

The discriminator and the generator are involved in a min-max game with the value function $V(D,G)$ as given below:
\vspace{-1mm}
\begin{equation*}
\max_{D}\min_{G} V(D,G) = \mathop{\mathbb{E}}_{x\sim P_{\text{Data}}(x)}[\log(D(x)] + \mathop{\mathbb{E}}_{z\sim P_Z(z)}[1-\log(D(G(z)))]
\vspace{-1mm}
\end{equation*}
\vspace{0mm}
While optimizing the discriminator and the generator network, the following cost functions are used: 
\vspace{-1mm}
\begin{align*}
\begin{split}
J^{(D)}& = -\frac{1}{2}\mathbb{E}_{x\sim P_{\text{Data}}(x)}[\log(D(x)] -\frac{1}{2}\mathbb{E}_{z\sim P_Z(z)}[1-\log(D(G(z)))] \\
J^{(G)}& = -J^{(D)}
\end{split}
\vspace{-1mm}
\end{align*}
\vspace{0mm}
\noindent where $J^{(D)}$ and $J^{(G)}$ represent the costs corresponding to the discriminator and the generator respectively.
 
However in practice, $J^{(G)}=-\log(D(G(z)))$ is used as the generator objective function, as it avoids early saturation of the gradient loss~\cite{NIPS2014_5423}. Even with this objective function it is sometimes very difficult to train GANs as it may suffer from mode-collapse and non-convergence~\cite{arjovsky2017towards}. Hence, we use a recent variation of GAN called Wasserstein GAN~\cite{DBLP:conf/icml/ArjovskyCB17}, which is more stable and has been shown to perform better in practice.

\subsection{Wasserstein GAN (WGAN)}
\label{WGAN}
Essentially, variations of GANs are models that learn the real data distribution through the minimization of $f$-divergence. $f$-divergence is a type of metric family, which measures the distance or divergence between the real and fake data distribution. The original GAN paper~\cite{NIPS2014_5423} used minimization of Jenson-Shannon (JS) divergence. It is often difficult to train a GAN using JS divergence as it is not guaranteed to be continuous and differentiable everywhere~\cite{DBLP:conf/icml/ArjovskyCB17}. In \cite{DBLP:conf/icml/ArjovskyCB17}, the authors propose the Earth Mover's distance as a better alternative to the conventional divergence based measures. We describe the WGAN in detail now.

Let $P_r$ be the real data distribution. We want to learn the generator distribution $P_g$, which tries to approximate $P_r$. We approximate the distribution using a neural network based function approximator $g_\theta(z)$, where $\theta$ is a learnable parameter, such that $P_g = g_\theta(z)$. In WGAN, we try to minimize the following distance:
\begin{equation*}
W(P_r, P_g) = \inf_{\gamma \in \prod(P_r,P_g)} \mathbb{E}_{(x,y) \in \gamma}[\left\Vert x-y \right\Vert]
\end{equation*}
Unfortunately, computing the exact Wasserstein distance is intractable. The nearest approximation to this is a result from Kantorovich Rubinstein duality~\cite{rachev1990duality}, which shows that $W$ is equivalent to:
\begin{equation*}
W(P_r, P_g) = \sup_{{\left\Vert f \right\Vert}_L \leq 1} \mathbb{E}_{x \sim P_r}[f(x)] - \mathbb{E}_{x \sim P_g}[f(x)]
\end{equation*}
\noindent Where the supremum is taken over all the \textbf{1} - Lipschitz functions~\cite{DBLP:conf/icml/ArjovskyCB17} $f:\mathcal{X}\to\mathbb{R}$.
The final update equation is given by:
\begin{equation*}
\nabla_{\theta}W(P_r,P_g)  = \nabla_\theta(\mathbb{E}_{x \sim P_r}[f_w(x)] - \mathbb{E}_{z \sim p_Z(z)}[f_w(g_\theta(z))])
\end{equation*}
Here $f_w$, which approximates the mapping of the input to the output in the discriminator, is the optimal Lipschitz continuous function for the Wasserstein distance. We note that the final layer of the discriminator does not contain any non-linearity.

Now we describe the steps of WGAN. First, for a fixed $\theta$, we compute an approximation of Wasserstein distance by training $f_w$ until convergence. Using the optimal function $f_w$, we sample several $z \sim p_Z(z)$ and compute the gradient of $\theta$: $-\mathbb{E}_{z \sim p_Z(z)}[\nabla_\theta f_w(g_\theta(z))]$. Next, we update $\theta$ and repeat the process. Since the entire premise of using Wasserstein loss is based on the necessary condition that function $f_w$ is Lipschitz continuous, we are required to use weight clamping and hence weights have to lie within the compact space $[c,-c]$, where $c$ is a small constant. This problem of manual clamping has been alleviated in a follow-up paper~\cite{gulrajani2017improved}, which introduces a soft-version of the \textbf{1}-Lipschitz constraint by forcing a penalty on the gradient norm. The new losses corresponding to the discriminator and the generator are denoted as $J^{(D)}_W$ and $J^{(G)}_W$ respectively and are defined as:
\begin{align*}
\begin{split}
J^{(D)}_W &= -\mathop{\mathbb{E}}_{x \sim P_g}[D(x)] + \mathop{\mathbb{E}}_{x \sim P_r}[D(x)] - \lambda\mathop{\mathbb{E}}_{x \sim P_{x'}}[{\left\Vert \nabla_xD(x) \right\Vert}_2 - 1]^2\\
J^{(G)}_W &= -J^{(D)}_W
\end{split}
\end{align*}
Here $P_{x'}$ is a uniform distribution defined on a straight line formed between pairs of points which are being sampled from $P_r$ and $P_g$, respectively. This framework is incorporated into ecGAN and ec\textsuperscript{2}GAN to learn an accurate distribution of e-commerce orders.

\subsection{Order Representation}
\label{order_representation}

An e-commerce order is defined in terms of the following key elements: \{customer, product, price, date\}. We compute feature representations for each of these individual elements and concatenate them to obtain a dense, low-dimensional representation for the order. This order representation is eventually fed to the proposed ecGAN and ec\textsuperscript{2}GAN. In what follows, we describe how we compute the representations corresponding to each key element of an e-commerce order. We note that e-commerce websites often involve customers purchasing from a marketplace of sellers. In future work, we intend to expand our definition of orders to explicitly incorporate information pertaining to sellers as well.

\subsubsection{Product embeddings}
\label{product-embedding}
We use an unsupervised approach to represent each product in an e-commerce ecosystem using a dense and low-dimensional representation. The word2vec representations of all the words in a product title are (weighted-) averaged to create dense product embeddings. First, we train a word2vec~\cite{mikolov2013efficient} model using a corpus of titles and descriptions pertaining to 143 million randomly selected products. We obtain a vocabulary of around 1.4 million words along with their word2vec representations. Next, the same word2vec corpus is used to find the inverse document frequency (IDF) of all the words in the vocabulary. The word2vec representations corresponding to each word are multiplied with the corresponding IDF weights and are summed across all the words in the title. The representations are normalized by the total IDF score of all the words in the title. This approach brings similar products close to each other in the learned representation space. The embeddings created using this method are 128 dimensional and the values range between -1 and 1.

\begin{figure}[ht!]
    \centering
    \includegraphics[width=0.85\columnwidth]{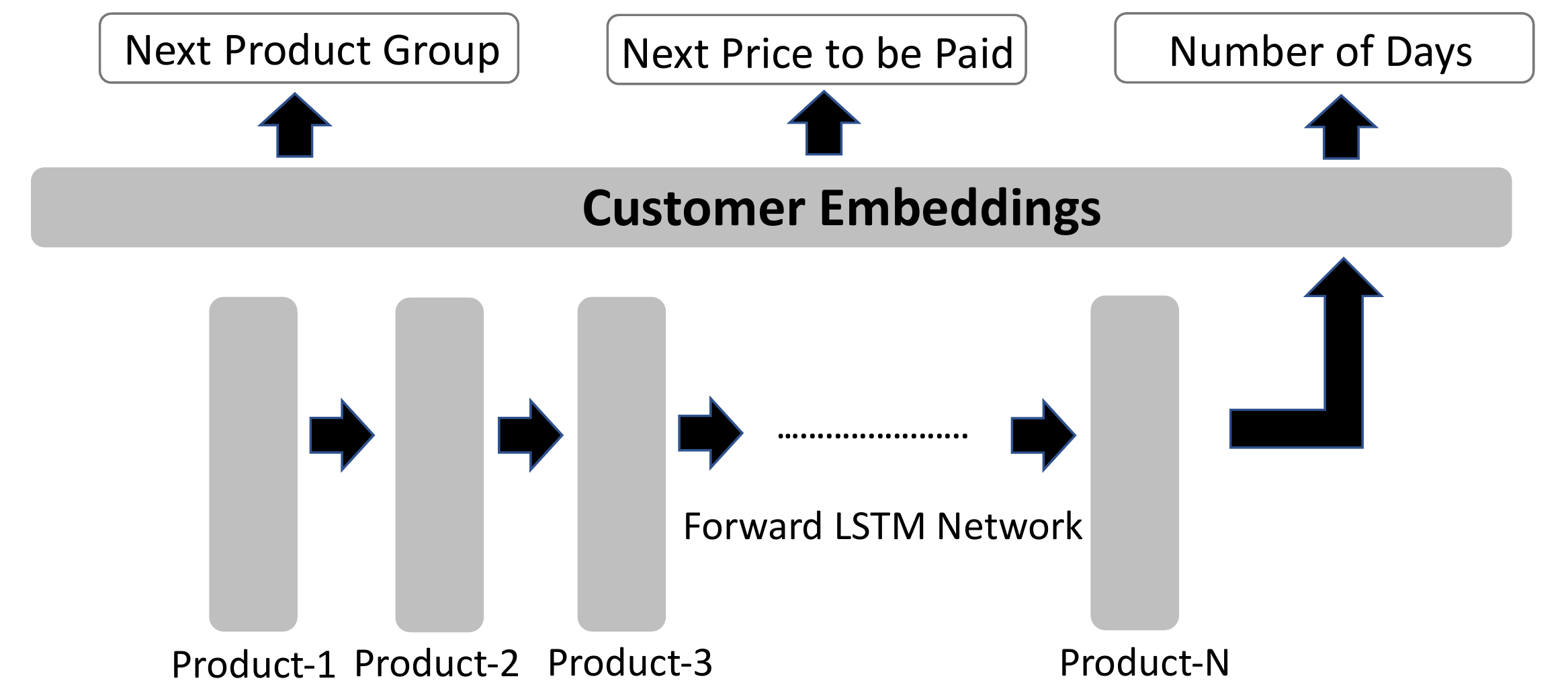}
    \caption{The Network for Customer Embeddings.}
    \label{cust2vec}
\end{figure}

\begin{figure*}[ht!]
    \centering
    \includegraphics[width=0.95\textwidth]{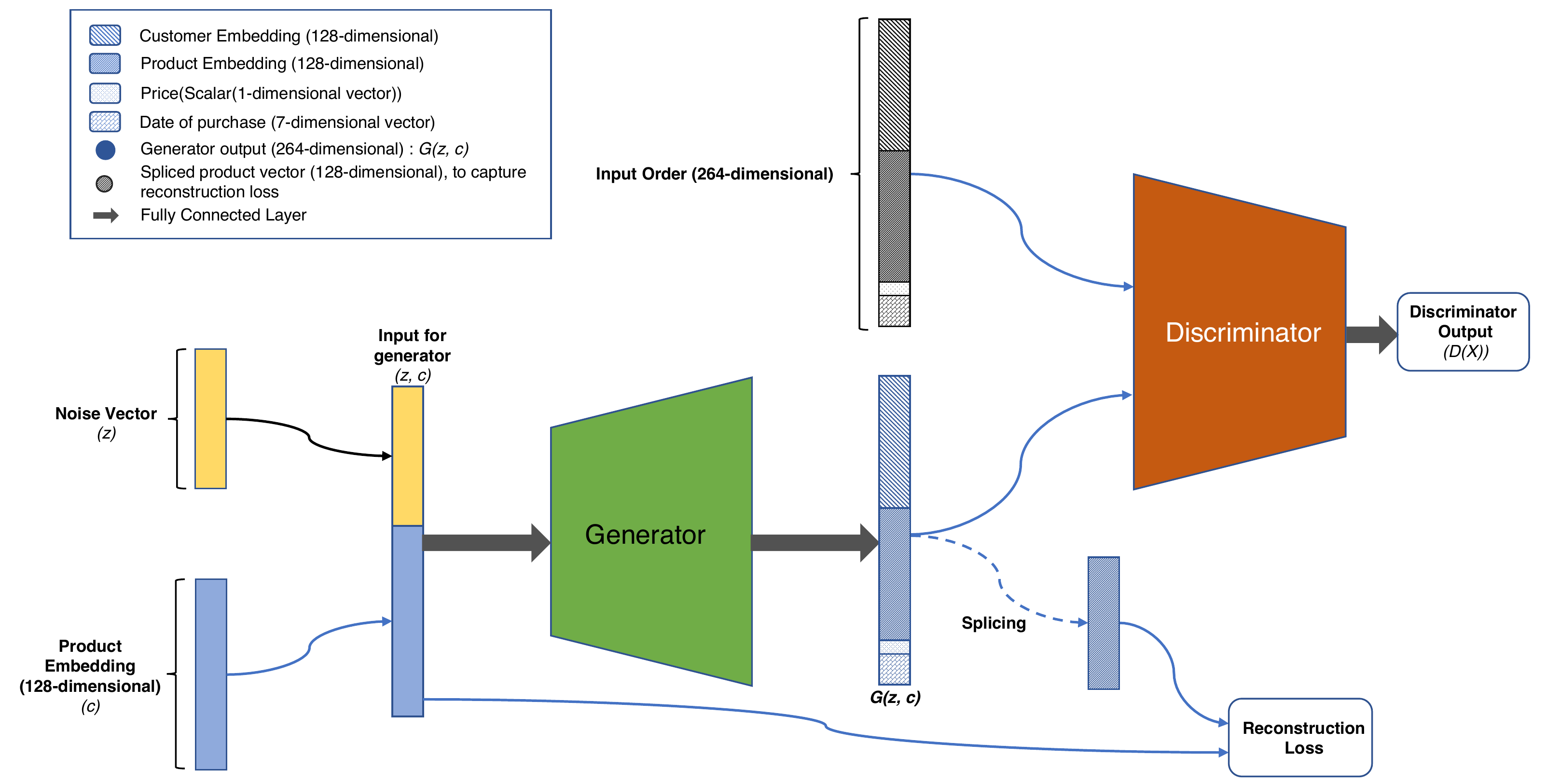}
    \caption{The architecture of the proposed ec\textsuperscript{2}GAN. The ecGAN architecture is same as the ec\textsuperscript{2}GAN architecture except that it does not feed the product embeddings to the generator and the generator does not optimize any reconstruction loss (best viewed in electronic copy).}
    \label{arch_1}
\end{figure*}

\subsubsection{Customer embeddings}
To represent each customer in an e-commerce system, we train a Discriminative Multi-task Recurrent Neural Network (RNN), where different signals pertaining to a customer's recent purchase history are explicitly encoded into her embedded representation (Figure \ref{cust2vec}). Each signal is captured by formulating a multi-class classification task. We use three different training tasks: (i) Predicting the next product group (such as clothes, food, furniture, baby-products etc.) purchased by a customer, (ii) Predicting how much price a customer will pay on the next purchase and (iii) Predicting after how many days the customer will purchase an item from the e-commerce company\footnote{We tried various training task combinations for a fixed embedding dimension and found this subset to be most effective on various customer classification tasks.}. The proposed architecture contains an RNN with LSTM cells as the input layer which takes the sequence of products (i.e., their representations as described in Section \ref{product-embedding}) purchased by a customer and creates a hidden representation. The hidden representation, which we refer to as ``customer embedding'', is fed into multiple classification units corresponding to the training tasks. The network is jointly trained with all the tasks in an end-to-end manner using alternating optimization. At each iteration of training, we randomly choose one of the tasks and optimize the network with respect to the loss corresponding to that task only. This approach brings two customers close in the semantic space if they have similar purchase history. Using this approach we represent each customer using a 128 dimensional dense and low-dimensional vector. The range of each feature is between -1 and 1.

\subsubsection{Price}
The prices of the products sold in an e-commerce system varies between a few dollars to tens of thousands of dollars. We first take logarithm of the prices to squash the effect of the large price values. We further normalize each log-transformed price between the range of -1 and 1.

\subsubsection{Date of Purchase}
Date of purchase of an order is represented as a 7-dimensional vector. The first component captures the difference between the current date and a pre-decided epoch. The next two components represent day of the month, followed by two components representing the day of the week and the last two components representing the month. The features have been carefully crafted to contain the information about circularity of days, i.e., Monday would be equally close to Wednesday as it would be to Saturday. We achieve this by projecting the possible days/months on the periphery of a unit circle and representing them using their sine and cosine components. Each of the features are normalized between -1 and 1.

We concatenate the 128 dimensional customer vector, 128 dimensional product vector, 1 dimensional price and 7 dimensional purchase date vector to obtain a 264 dimensional vector corresponding to each order. These vectors are dense, low-dimensional and semantically meaningful.

\begin{figure*}[ht!]
    \centering
    \includegraphics[width=0.80\linewidth]{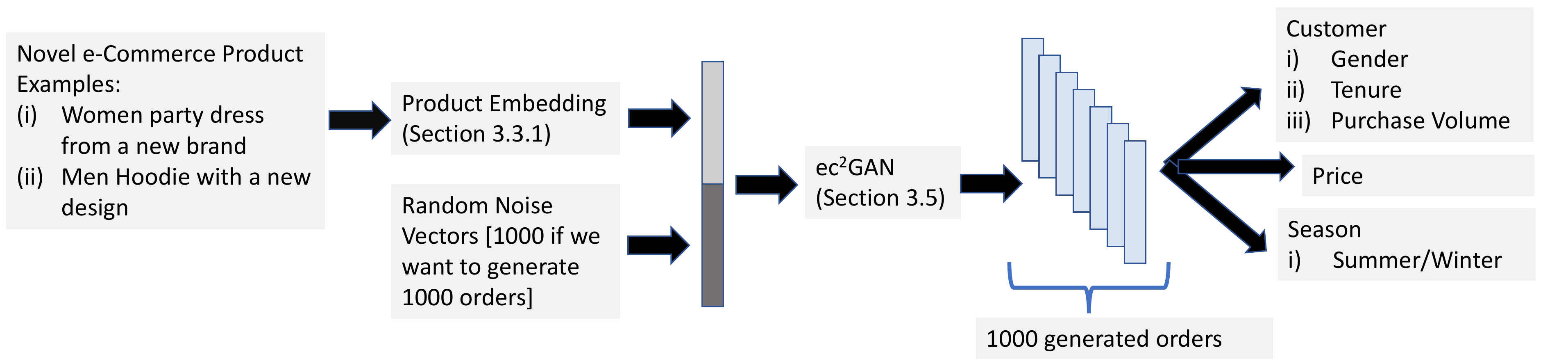}
    \caption{The Application Pipeline for ec\textsuperscript{2}GAN.}
    \label{ec2gan_application_pipeline}
\end{figure*}

\subsection{ecGAN}
\label{ecgan}
In this version, we use WGAN (Section \ref{WGAN}) to model the space of order representations. Let us assume that an order $\mathbf{O}_n\in\Phi$ is a tuple $\{\mathbf{C}_i, \mathbf{P}_j, p_k, \mathbf{D}_l\}$, where $\mathbf{C}_i$ is the representation of the i-th customer at that point in time when she makes this order/purchase, $\mathbf{P}_j$ is the product representation of the j-th product, $p_k$ is the price of the product and $\mathbf{D}_l$ represents the vector corresponding to the date on which the order was made. The discriminator in ecGAN is fed with the set of orders $\Phi$ and it tries to discriminate them from the order set $\tilde{\Phi}$, which is generated by the ecGAN generator.

The generator, which is a fully connected network with two hidden layers and ReLU~\cite{nair2010rectified} non-linearity at the end of each layer, maps the noise vectors ($z$) to feasible orders ($\tilde{\mathbf{O}_n}$). The 264-dimensional fake orders ($\tilde{\mathbf{O}_n}$) from the generator are compared with the 264 dimensional real orders ($\mathbf{O}_n$) in the discriminator. The discriminator, which is also a fully connected network with two hidden layers and ReLU~\cite{nair2010rectified} non-linearity, maps the order inputs to real or fake labels. Post-convergence, the orders which are generated by the ecGAN generator are evaluated in several qualitative ways as described in Section \ref{ecgan_results}.

\subsection{ec\textsuperscript{2}GAN}
\label{ec2gan}


In this version, we want the generator to generate orders which are conditioned on a particular product. The block diagram of the proposed ec\textsuperscript{2}GAN is shown in Figure \ref{arch_1}. We make several changes to ecGAN and propose an ecommerce-conditional-GAN or ec\textsuperscript{2}GAN which can be conditioned on any product and orders involving that product could be generated. To this end, we make a couple of key modifications to the basic ecGAN architecture:
\begin{itemize}[noitemsep, leftmargin=*]
\item In the generator component, we feed a vector $z^{\prime}$, which is the concatenation of random noise vector $z$ and the product representation $\mathbf{P}_j$, i.e., $z^{\prime}=[z,\mathbf{P}_j]$. This ensures that during test time, when we want to generate a set of feasible orders, we can condition the network on any product of our choice.
\item We also add a reconstruction loss component to the generative loss. The reconstruction loss, denoted as $J^{(R)}$, enforces the product components of the generated orders to be exactly same as the product representation which we conditioned on. We define the reconstruction loss as the Euclidean distance between the actual product vector $\mathbf{P}_j$, which is the input to the generator, and the generated product vector $\tilde{\mathbf{P}_j}$, i.e., $J^{(R)}=||\mathbf{P}_j-\tilde{\mathbf{P}_j}||$. The new generator loss in the ec\textsuperscript{2}GAN is defined as $\alpha J^{(G)}_W+(1-\alpha)J^{(R)}$, where $\alpha$ is a tunable parameter.
\end{itemize}

The generator and discriminator network architecture in ec\textsuperscript{2}GAN remains exactly same as that of ecGAN.

\subsection{Application Pipeline for ec\textsuperscript{2}GAN}

In this section, we describe how the proposed ec\textsuperscript{2}GAN can be applied to a novel e-commerce product in real-world scenarios. The application pipeline is provided in Figure \ref{ec2gan_application_pipeline}. The steps are described below:
\begin{itemize}[noitemsep, leftmargin=*]
\item First, we obtain an embedding for a product using the proposed approach in Section \ref{product-embedding}.
\item The product embedding is concatenated with random noise vectors for feeding them as input to the ec\textsuperscript{2}GAN network. If we want to generate 1000 orders, we sample 1000 different random noise vectors, concatenate each of them with the product embedding and the concatenated vectors are provided as input to the ec\textsuperscript{2}GAN. The ec\textsuperscript{2}GAN outputs 1000 different orders. However, for a particular product, we can generate any number of plausible orders. 
\item The generated orders are classified to determine the customer preference (gender, tenure and purchase volume), price and seasonal demands of a product. Further details are provided in Section \ref{cust_char}, \ref{price_char} and \ref{season_char}.
\end{itemize}

\section{Experimental Results}

In this section, we discuss the experimental results. In Section \ref{dataset}, we provide the details of the dataset used in the experiments. To successfully train the GANs used in this work, we had to resort to several non-standard strategies and improvisations, which are documented in Section \ref{train_facts}. Next, in Section \ref{ecgan_results}, we propose several qualitative measures to evaluate the ecGAN. Finally, the quantitative results on ec\textsuperscript{2}GAN are presented in Section \ref{ec2gan_results}.
 
\subsection{Dataset}
\label{dataset}
In this work, we use the products from the apparel category for model training and evaluation. We randomly choose $5$ million orders made over the last one year in an e-commerce company to train
the proposed models. Once we obtain the \{customer, product, price, date\} tuple corresponding to each order, we get their corresponding 264-dimensional representations using the proposed order representation approach (Section \ref{order_representation}) and train our models.

\subsection{What did we learn?}
\label{train_facts}
We tried various tricks and parameters to obtain the stable e-commerce GAN variations. We document them below for the benefit of interested readers:

\begin{figure*}[ht!]
    \centering
    \begin{subfigure}[b]{0.33\linewidth}
        \includegraphics[width=\linewidth]{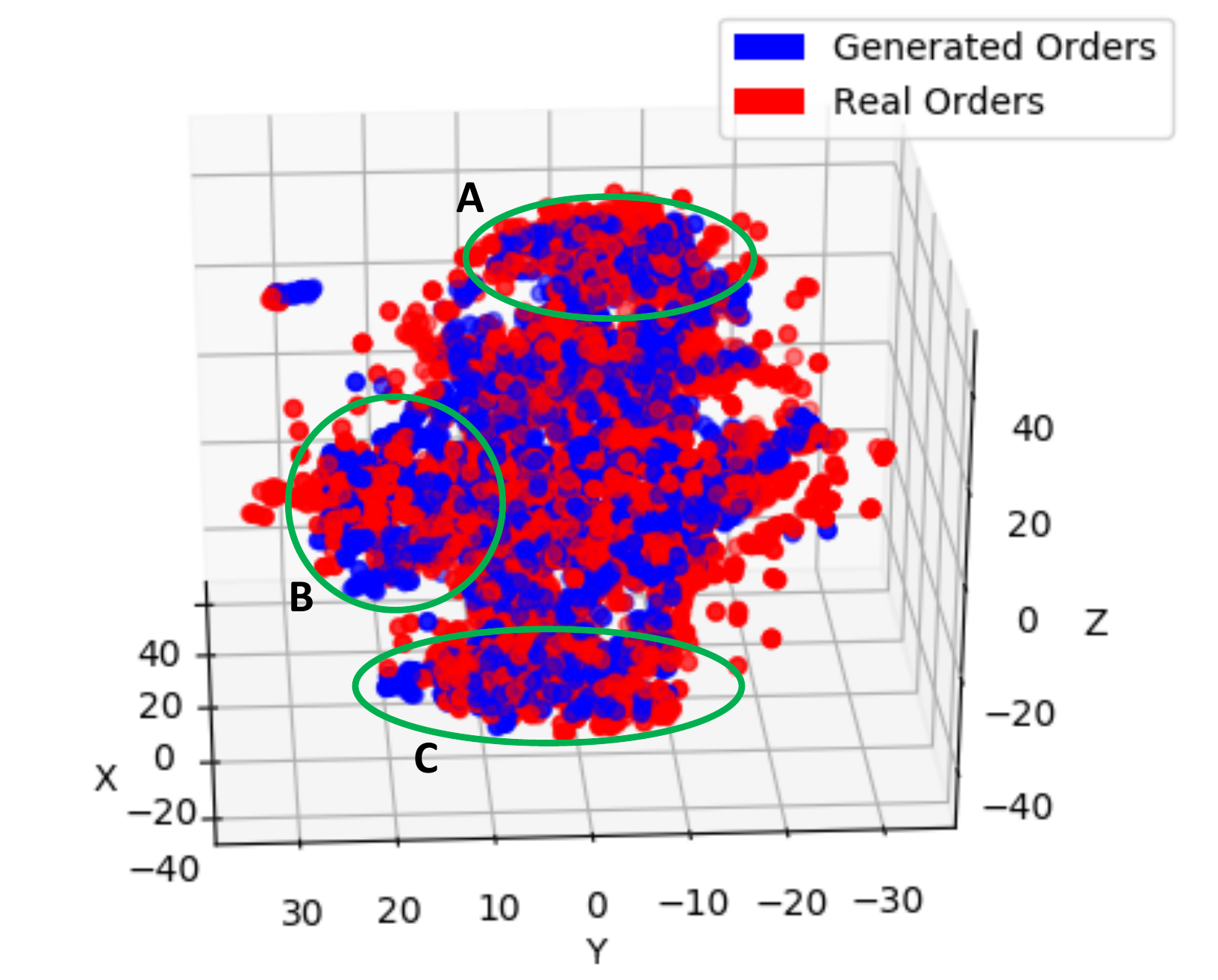}
        \caption{3D t-SNE: Perspective 1}
        \label{3d1}
    \end{subfigure}
    ~ 
    \begin{subfigure}[b]{0.33\linewidth}
        \includegraphics[width=\linewidth]{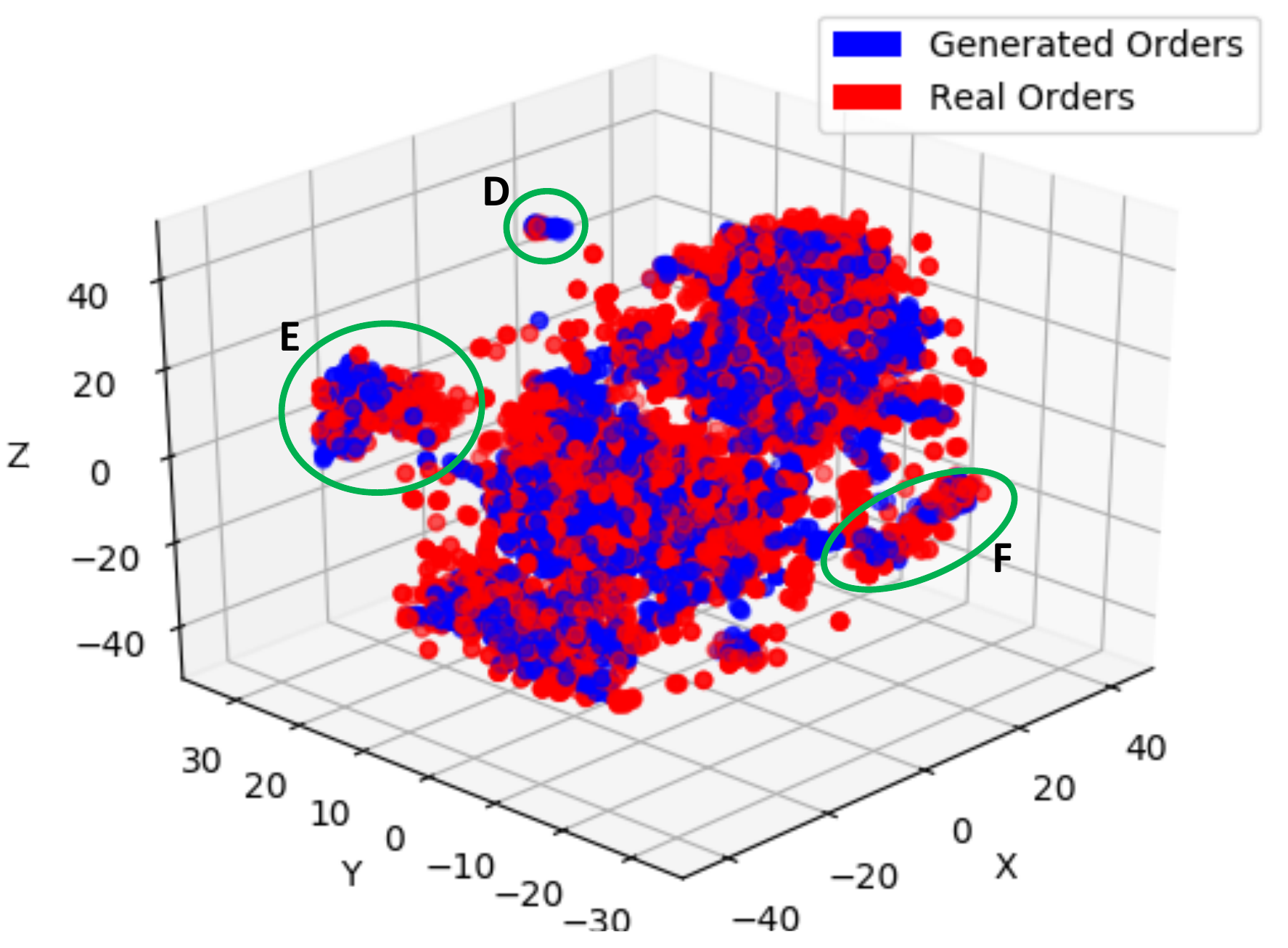}
        \caption{3D t-SNE: Perspective 2}
        \label{3d2}
    \end{subfigure}
    \begin{subfigure}[b]{0.33\linewidth}
        \includegraphics[width=\linewidth]{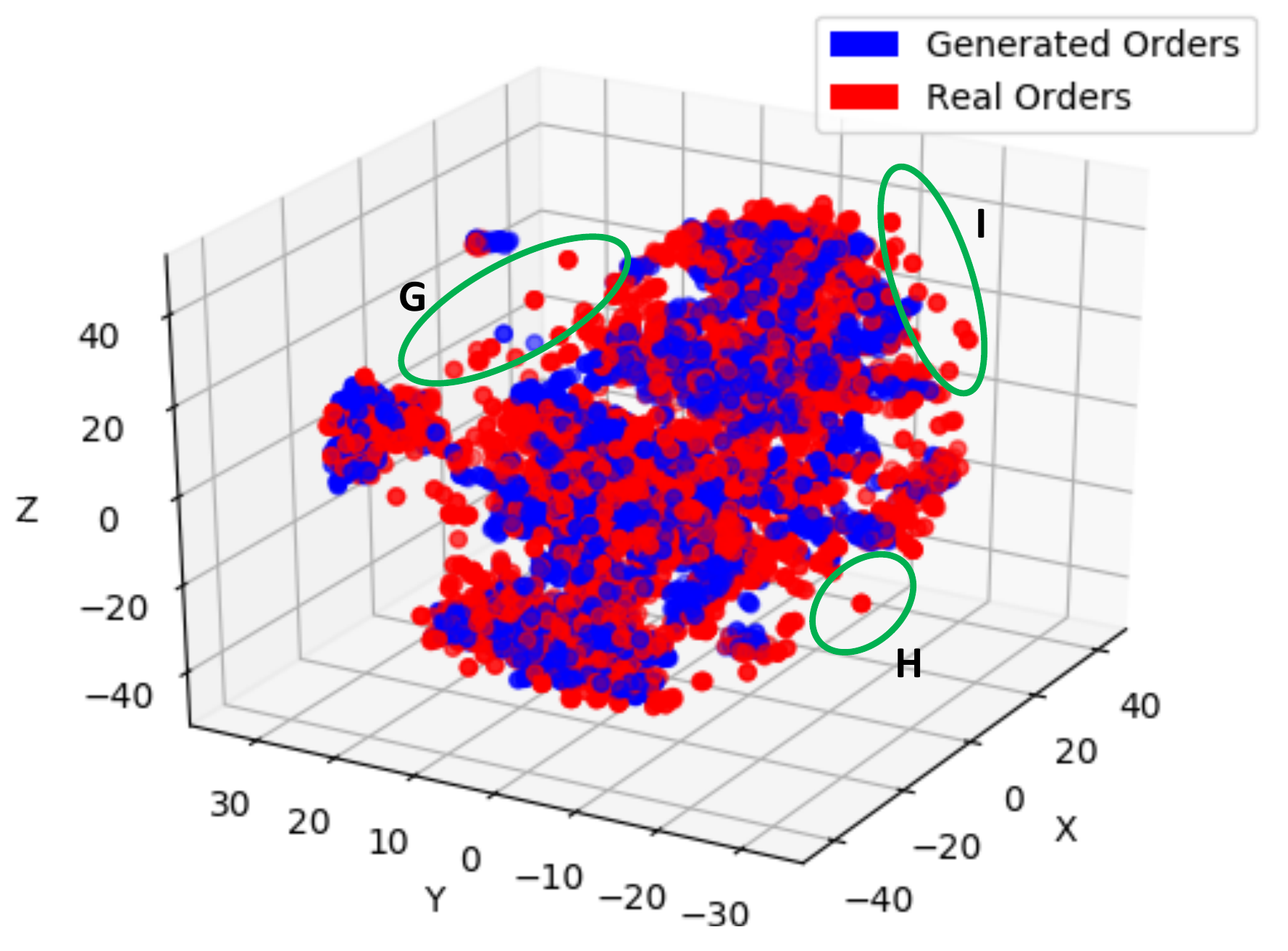}
        \caption{3D t-SNE: Perspective 3}
        \label{3d3}
    \end{subfigure}
    \caption{Different perspective of \textit{orders} in a projected 3-D space (t-SNE). Blue represents generated \textit{orders}, whereas red denotes real \textit{orders} (best viewed in electronic copy).}\label{3dtsne}
\end{figure*}

\begin{itemize}[noitemsep, leftmargin=*]
\item We use multilayer fully-connected networks for both the discriminator and the generator. To stabilize the GAN trainings, we experimented with varying the number of discriminator layers, while keeping the number of generator layers constant. We observe that if we increase the number of layers in the discriminator as compared to the generator, the discriminator becomes too strong in the initial iterations itself leading to early convergence of both the discriminator and the generator. On the other hand, if we reduce the number of layers in the discriminator, the learned generator fails to effectively model the real data distribution. After experimenting with various number of layers, we choose two layers in both the discriminator and the generator. 
\item We use ReLU activation in both the generator and the discriminator. However, in the last layer of the generator, we use tanh activation, as the generated data should lie between -1 and 1.
\item While keeping the number of layers in the generator fixed at two, we tried various input noise dimensions: $\{32, 64, 96, 128\}$. We found that 96-dimensional noise vector to the generator gives the most stable and accurate e-commerce GAN.
\item We used the Logistic Regression accuracy to track the learning of the proposed GANs and tune the parameters. We randomly sample 10K real orders and generate 10K fake orders using the generator. We assign label of 1 to the real data points and label of 0 to the fake data points. We combine both these datasets and randomly permute them to create a dataset of size 20K. The data is randomly split into training part (80\% of the data) and test part (20\% of the data). The accuracy on the test dataset is used to track the quality of the generator. If the Logistic Regression gives an accuracy of 100\%, the generator has not learned any useful representation and the discriminator is able to easily distinguish between the real and fake data. However, an accuracy of 50\% by the Logistic Regression indicates that the generator is able to mimic the real data quite well. Logistic Regression provides a fast yet simple approach to track the learning of GANs.
\end{itemize}

A few architectural details about the proposed ecGAN and ec\textsuperscript{2}GAN are enumerated below:
\begin{enumerate}[noitemsep, leftmargin=*]
\item Generator (two hidden layered network): $\mathbf{96}$ (in ecGAN) and $\mathbf{96+128=224}$ (in ec\textsuperscript{2}GAN) $\to \mathbf{64} \to \mathbf{128} \to \mathbf{264}$ 
\item Discriminator (two hidden layered network): $\mathbf{264} \to \mathbf{128} \to \mathbf{64} \to \mathbf{1}$
\item $\alpha: \mathbf{0.75}$
\item $\beta$ (used in the Adam optimizer): $\mathbf{0.6}$
\item Ratio of Number of times discriminator is trained as compared to generator: $\mathbf{5:1}$
\item Minibatch size: $\mathbf{128}$
\item Number of epochs for convergence (5 million orders): $\mathbf{15}$
\end{enumerate}

\subsection{ecGAN Qualitative Analysis}
\label{ecgan_results}
Although GANs have been successful in various applications, evaluation of GANs is a difficult task~\cite{theis2015note, wu2016quantitative, quantitative_eval_gan}. When GANs are used with image data or music data, it is possible to view the generated images or listen to the generated musics and qualitatively check if the GAN has effectively learned the real data distribution~\cite{denton2015deep, radford2015unsupervised, mogren2016c}. However, in various scenarios such as ours, the generated data is not human perceptible and hence can not be directly evaluated. We propose three different approaches to qualitatively study the effectiveness of the trained GANs and demonstrate that the order generator has learned to generate plausible e-commerce orders.
\begin{itemize}[noitemsep, leftmargin=*]
\item {\bf t-SNE~\cite{van2008visualizing}:} t-Distributed Stochastic Neighbor Embedding (t-SNE) is a technique for dimensionality reduction that is particularly well suited for the visualization of high-dimensional datasets. The technique can be implemented via Barnes-Hut approximations, which allows it to be applied on large real-world datasets. We randomly choose 10K real orders and generate 10K orders using ecGAN. We mix all the data and perform t-SNE and project them to a 3-dimensional space. The projected space is explored from various perspectives and are shown in Figure \ref{3dtsne}. We make three key observations: (a) the real orders and the generated orders are mixed well with each other throughout the projected space (e.g., region A, B and C in Figure \ref{3d1}), (b) the generator has learned to generate data from all the modes in the real order distribution (e.g., region D, E and F in Figure \ref{3d2}) and (c) the generator has learned to ignore the outliers or unlikely orders in the real order-set (e.g., region G, H and I in Figure \ref{3d3}). Overall, ecGAN is able to generate orders that are quite similar to the real order distribution.
\item {\bf Feature Correlation:} In this study, we compute the feature correlation between the real orders and the generated orders. There are total 264 features in an order. We randomly select three features $\{f_1,f_2,f_3\}$ from the feature set. We check if the correlation coefficient between $f_1$ and $f_2$ is more than the correlation coefficient between feature $f_1$ and $f_3$. For each triplet, we check if the feature relationship in the real data agrees with the feature relationship in the generated data. We randomly choose 100K triplets and determine the fraction of the times they agree. We observe that $77\%$ of the times, the generated data match with the real data (whereas the baseline will be $50\%$). This also demonstrates that the generator has learned to generate orders which are similar to the real orders to an appreciable extent.
\begin{figure}[h]
    \centering
	\includegraphics[width=\linewidth]{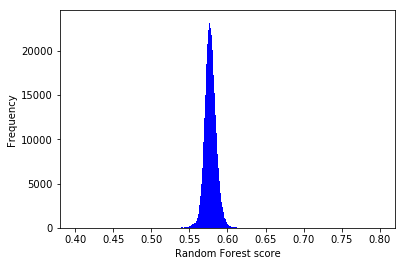}
    \caption{Histogram of data distribution scores in random forest leaves.}
    \label{hists}
\end{figure}

\item {\bf Data distribution in Random Forest Leaves:} We propose an indirect method to estimate the similarity between the real orders and the fake orders. We train a random forest classifier and look at the distribution of the real and fake orders at each leaf node of the forest. We randomly sampled 1 million real orders, generated 1 million fake orders using ecGAN and mixed them to create a set of 2 million orders. We label them as class 1. Next, we randomly permuted the columns of these orders and generate a random dataset of 2 million orders. This dataset is labeled as class 2. We train a binary classification model to separate class 1 from class 2. The random forest has 100 trees with maximum depth of 5 (32 leaf nodes in each tree, i.e., 32 clusters of the input data). Next, we trace each of the 1 million real orders and 1 million fake orders through each of the 100 trees and note down the nodes they traveled through. We compute for each leaf in each tree the “real-order-ratio”, which is defined as the fraction of the real orders out of all the orders that end up in that node. Then for each order, we average the “real-order-ratio” of the leaf nodes they end up in across all the 100 trees. If in each node, the real data and fake data are mixed up well, these scores should be close to 0.5 and should indicate that the generated orders are similar to the actual orders. We plot the histogram of the scores of all the real orders and plot them in Figure \ref{hists}. This distribution (centered around 0.57) establishes that the real data and fake data are quite inseparable .

\end{itemize}

\subsection{ec\textsuperscript{2}GAN Quantitative Analysis}
\label{ec2gan_results}

As discussed in the earlier sections, GAN evaluation is difficult. We propose an approach to quantitatively evaluate GANs. The evaluation technique is designed with e-commerce in mind but should be applicable to any other domain such as finance, transportation, health-care and sports. In ec\textsuperscript{2}GAN, a product representation is fed along with the noise vector to the generator and orders which involve the input product are generated. When a new product is introduced into an e-commerce system, it is extremely hard to determine the set of customers who would be interested in this product, at how much price the product will be sold and when the product will have high demand. Robust forecasting of these would enable e-commerce companies to better manage their inventories and adopt proactive supply chain optimizations. In this section, we discuss given a brand new e-commerce product, how we can characterize various components (customer demographics, price, date or seasonality) of the generated e-commerce orders. We propose an evaluation metric called Relative Similarity Measure or RSM which lets us quantitatively compare the generated orders and the real orders with respect to each characteristic. We note that the goal of this paper is not to produce state-of-the-art results in customer characterization, price estimation or demand prediction, where highly specialized solutions can be engineered for each of these problems independently. We apply ec\textsuperscript{2}GAN to these problems to demonstrate a quantitative way of evaluation of the proposed GAN technique, which is difficult to do otherwise. We compare ec\textsuperscript{2}GAN with a baseline order generation approach based on Conditional Variational Autoencoder (C-VAE)~\cite{C-VAE} and describe the details of the baseline approach in Section \ref{baseline_approach}. The data which are used in the following evaluation experiments are completely disjoint from the dataset, which we used to train our model.

\subsubsection{Baseline Order Generation Approach}
\label{baseline_approach}

We compare the proposed approach with a baseline approach that uses Conditional Variational Autoencoders~\cite{C-VAE} for order generation. This approach is an extension of Variational Autoencoders~\cite{VAE}. However, the generation process can be conditioned on various kinds of signals such as labels, images etc., such that the generated samples are restricted by the input signal. For example, in \cite{C-VAE}, the authors propose to predict the future such as what will move in a scene, where it will travel, and how it will deform given any static image. In this work, we use the C-VAE for order generation given a particular product. The same dataset is used for training both the proposed ec\textsuperscript{2}GAN and the baseline C-VAE. 
\subsubsection{Relative Similarity Measure (RSM)}

We propose the RSM to measure the relative similarity between the generated orders and the real orders. Let us assume that each product $p_i$ can be assigned a propensity score of $s_i^t$ by the ground-truth order statistics and a propensity score of $s_i^g$ by the generated order statistics corresponding to each intrinsic characteristic. In this paper, we consider the following hierarchical structure of the intrinsic characteristics:
\begin{itemize}[noitemsep,leftmargin=*]
\item {\bf Customer}
\begin{itemize}
\item Gender
\begin{itemize}
\item Female
\item Male
\end{itemize}
\item Tenure
\begin{itemize}
\item High-tenured (more than five years)
\item Medium-tenured (between two and five years)
\end{itemize}
\item Purchase Volume
\begin{itemize}
\item High Purchasers
\item Average Purchasers
\end{itemize}
\end{itemize}
\item {\bf Price}
\item {\bf Seasonal Demand}
\begin{itemize}
\item Summer (May, June, July, August)
\item Winter (November, December, January, February)
\end{itemize}
\end{itemize}

The propensity scores are computed for each of the leaf nodes in the above structure, e.g., female, male, high-tenured, medium-tenured, price etc. The details about computing the scores are provided in Section \ref{cust_char}, \ref{price_char} and \ref{season_char}. The propensity scores enforce an ordering of all the products with respect to both the ground-truth orders and the generated orders. We determine an agreement between these orderings using the following approach. For any two products $p_i$ and $p_j$, if $s_i^t\geq s_j^t$ and $s_i^g\geq s_j^g$ or $s_i^t<s_j^t$ and $s_i^g<s_j^g$, we count that as an agreement between the ground-truth orders and the generated orders. We randomly sample $N$ ($10000$ in this paper) pairs of products, find the percentage of the times the generated data and ground-truth data are concordant with each other and define that as the Relative Similarity Measure or RSM. A random baseline, where the scores are assigned randomly to each product will produce an RSM score of 50\%. 

\subsubsection{Customer Characterization}
\label{cust_char}
When a new product is launched, any e-commerce company will be interested in various characterization of the customers, who are likely to buy this product. The characterization could be based on gender, age, tenure, purchase volume etc. Once we know the characteristics of the customers, we could target relevant customers with specific deals and recommendations involving this particular product. Using ec\textsuperscript{2}GAN, we could generate a set of orders given a product and use the corresponding customer components for various characteristic analysis. In this paper, we analyze the gender, tenure and purchase volume of prospective customers of a product. We follow the steps below to obtain the propensity score corresponding to a customer related intrinsic characteristic:
\begin{enumerate}[noitemsep,leftmargin=*]
\item First, using internal survey and historical data we train classifiers which can predict the gender, purchase volume and tenure of a customer, given its representation. These classifiers are used to map each customer to one of the categories corresponding to each characteristic. For example, corresponding to gender, the customers are mapped to either female or male. 
\item For a product $p_i$, we obtain the historical purchase history for a year, i.e., we know the set of customers (say, $\{C_{i}\}$)  who purchased this product over the last one year. We take the female (gender) characteristic to explain the method of obtaining a propensity score. The customers $\{C_{i}\}$ are fed to the trained gender classifier (in Step 1), to determine the fraction of female customers who purchased a product. This gives us the ground-truth propensity score $s_i^t$ for the i-th product and for the female characteristic.
\item Each product $p_i$'s representation is also fed to the proposed ec\textsuperscript{2}GAN to generate 1000 fake customers (say, $\{\tilde{C_{i}}\}$). The generated customers are also fed to the trained classifiers (in Step 1), to determine the fraction of female customers who purchased the product. This provides us the generated orders based propensity score $s_i^g$ for the i-th product and for the female characteristic.
\item The propensity scores $s_i^t$ and $s_i^g$ from the ground-truth data and the generated data respectively are used to compute an RSM score corresponding to the female characteristic.
\item We perform this evaluation for all of the customer related characteristics and report the results in Table \ref{ec2ganperf}. We obtain the RSM scores of $81.08\%$ and $82.20\%$ for the female and male characteristic respectively. Similarly, with average purchase volume, high purchase volume, medium-tenured and high tenured, ec\textsuperscript{2}GAN agrees with the ground truth 83.94\%, 87.25\%, 60.58\% and 73.64\% of the times respectively. These results demonstrate that the proposed ec\textsuperscript{2}GAN is effective in generating orders (customers in this case), which are quite similar to the real orders placed on an e-commerce website.
\end{enumerate}

\begin{figure*}[ht!]
    \centering
    \includegraphics[width=0.80\linewidth]{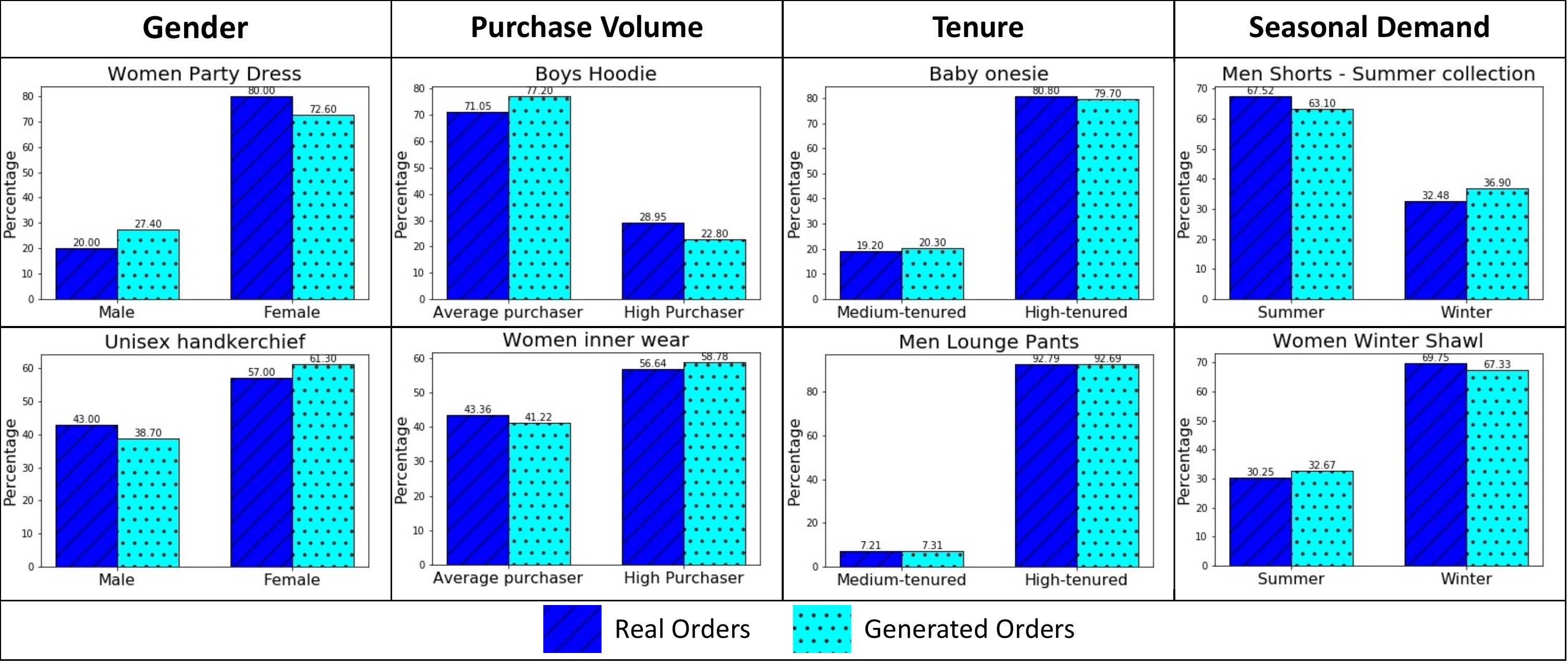}
    \caption{Distribution Comparison between  ec\textsuperscript{2}GAN and ground-truth orders. The plots are normalized, i.e., if x denotes \% of winter purchases and y denotes \% of summer purchases, we plot $\frac{x}{x+y}$ and $\frac{y}{x+y}$ respectively.}
    \label{dist_comp}
\vspace{-2mm}
\end{figure*}
\vspace{-2mm}

\vspace{-1mm}
\begin{table}[h]
\vspace{-1mm}
\caption{Customer Characterization (RSM scores reported).}
\vspace{0mm}
\begin{tabular}{|c|c|c|c|}\hline
\multicolumn{2}{|c|}{\textbf{Intrinsic Characteristics}}  & \textbf{ec\textsuperscript{2}GAN} &\textbf{C-VAE} \\ \cline{1-4}
  \multirow{2}{*}{Gender} & Female & {\bf 81.08\%} & 66.84\% \\ \cline{2-4}
  &  Male & {\bf 82.20\%} &  65.52\% \\ \cline{1-4}
  \multirow{2}{*}{Purchase Volume} & Average purchasers & {\bf 83.94\%} & 64.42\%  \\ \cline{2-4}
  &  High purchasers & {\bf 87.25\%} & 64.95\% \\ \cline{1-4}
\multirow{2}{*}{Tenure} & Medium-tenured & 60.58\% & {\bf 66.67\%}  \\ \cline{2-4}
  &   High-tenured & {\bf 73.64\%} & 65.7\% \\ \cline{1-4}
\end{tabular}
\label{ec2ganperf}
\vspace{-2mm}
\end{table}
\vspace{-2mm}

\subsubsection{Price Characterization}
\label{price_char}
When a new product is launched in an e-commerce website, accurate estimation of the price is important. The estimated price can be used to guide the sellers who might not always be aware of the actual worth of a product. Also, inaccurate catalog prices should be detected to avoid customer dissatisfaction. We can use ec\textsuperscript{2}GAN to estimate the price of a product and compare that with the ground-truth price. We find all the orders corresponding to a product over the last one year. The prices are averaged to obtain the ground-truth propensity score $s_i^t$. A product vector along with various noise vectors are fed to the proposed ec\textsuperscript{2}GAN to generate plausible orders of the input product. We generate 1000 orders corresponding to each product. The price component of the order vectors are extracted and averaged to determine the propensity score $s_i^g$. This score indicates the propensity of a product to be of higher price. The propensity scores are used to compute an RSM value corresponding to the price and the results are reported in Table \ref{ec2ganperf_price}. 

\vspace{-1mm}
\begin{table}[h]
\vspace{-1mm}
\caption{Price Characterization (RSM scores reported).}
\vspace{-1mm}
\begin{tabular}{|c|c|c|}\hline
\textbf{Characteristic} & \textbf{ec\textsuperscript{2}GAN} & \textbf{C-VAE} \\ \cline{1-3}
Price & 81.39\% & {\bf 86.51\%} \\ \cline{1-3}
\end{tabular}
\label{ec2ganperf_price}
\vspace{-2mm}
\end{table}
\vspace{-2mm}

\subsubsection{Seasonal Demand Characterization}
\label{season_char}

For a new e-commerce product, understanding its demand across seasons is critical for stock-keeping and inventory management. Using ec\textsuperscript{2}GAN, we look at the seasonal distributions of the generated orders. Specifically, for a particular product, we look at the propensity scores of it being sold in the summer season (May, June, July, August) and the winter season (November, December, January, February). For each product, we look at the one year purchase history and determine the propensity scores ($s_i^t$) corresponding to the summer and winter season. For each product, we also generate 1000 orders and based on the distribution of the months, we assign a summer and winter propensity score ($s_i^g$) to the product. The propensity scores are used to compute the RSM values corresponding to the summer and winter season (Table \ref{ec2ganperf_demand}). 

\vspace{-1mm}
\begin{table}[h]
\vspace{-1mm}
\caption{Seasonal Demand Characterization (RSM scores reported).}
\vspace{0mm}
\begin{tabular}{|c|c|c|}\hline
\textbf{Intrinsic Characteristics} & \textbf{ec\textsuperscript{2}GAN} & \textbf{C-VAE}\\\hline
Summer & {\bf 71.43\%} & 71.28\% \\\hline
Winter & {\bf 72.22\%} & 66.37\%  \\\hline 
\end{tabular}
\label{ec2ganperf_demand}
\vspace{-2mm}
\end{table}
\vspace{-2mm}

\subsubsection{Discussion}

In this section, we discuss the results obtained in Section \ref{cust_char}-\ref{season_char}. We compare the proposed ec\textsuperscript{2}GAN with the baseline C-VAE (Section \ref{baseline_approach}) for nine different use-cases (Table \ref{ec2ganperf}, \ref{ec2ganperf_price} and \ref{ec2ganperf_demand}). The proposed method is better than the baseline in seven out of the nine use-cases. The range of absolute improvement varied from minimum of 0.15\% (summer demand prediction) to maximum of 22\% (high-purchasing customer prediction). Overall, it is clear that the proposed approach is significantly better than the baseline approach. In general, we observe that the distribution of orders generated by the baseline C-VAE are peaky in nature, i.e., the variance is low. As a result of this, the generated orders often fail to capture the full spectrum of plausible orders corresponding to a particular product. However, ec\textsuperscript{2}GAN is much more effective in capturing the plausible order distribution of a particular product, which drives the better performance in customer characterization and seasonal demand prediction. We observe that, for price prediction, C-VAE performs better than ec\textsuperscript{2}GAN. Perhaps this can be attributed to the fact that the price distribution of most products have typically low variance, as the same product won't be sold at a wide-range of prices in an e-commerce website.  

\subsection{ec\textsuperscript{2}GAN: Distribution Comparison}
We also plot the normalized gender, tenure, purchase volume and seasonal demand distribution of the generated orders and the ground-truth orders for a few products in Figure \ref{dist_comp}. We observe that in almost all of these scenarios, the proposed ec\textsuperscript{2}GAN has been effective in emulating the ground-truth distribution. For example, given a product with title ``Women Party Dress'', the generated orders indicate that it will be purchased by female customers 73\% of the times, whereas male customers could buy this product 27\% times \footnote{A female oriented product is typically bought by males for gifts or family members.}. Similarly, ec\textsuperscript{2}GAN predicts that ``Men Shorts'' will be purchased during summer months 63\% of the times, whereas ``Women Winter Shawl'' will be purchased 67\% of the times during winter. Although it is often possible to determine the gender or season from the details of a product, the tenure or purchase volume of the prospective customers are not immediately inferable from the product details. In such scenarios, generating the possible orders of a product using ec\textsuperscript{2}GAN seems to be a natural way for order characterization. From Figure \ref{dist_comp}, it is clear that ec\textsuperscript{2}GAN has been able to mimic the real purchase volume and tenure data distribution in various products such as ``Boys Hoodie'', ``Women inner wear'', ``Baby onesie'' and ``Men Lounge Pants''. These results indicate the effectiveness of the proposed ec\textsuperscript{2}GAN. We also note that all these analysis would have been fraught with inaccuracies had we been plagued with issues of mode-collapse. The fact that the characteristics measured from the generated samples closely mimic the characteristics of the real samples, indicates the efficacy of our trained model.

\section{Conclusion}
In this paper, we propose two novel variations of Generative Adversarial Networks for e-commerce, ecGAN and ec\textsuperscript{2}GAN. We represent each order placed through an e-commerce website using a dense and low-dimensional representation. The proposed ecGAN learns to generate orders which are similar to the real orders. We evaluate the ecGAN qualitatively in three different ways and demonstrate its effectiveness. ec\textsuperscript{2}GAN is applied to generate orders, which will be placed for a new product that has been just introduced into an e-commerce system. The generated orders are characterized in various ways. We characterize the customers who would be interested in buying this product, the actual price of the product and seasonal demand of the product. We perform thorough quantitative analysis and demonstrate that ec\textsuperscript{2}GAN performs significantly better than the baseline on these tasks. We have proposed various techniques to qualitatively and quantitatively evaluate GANs in a novel application domain such as e-commerce. These technique should inspire successful GAN application and evaluation in various other domains such as healthcare, transportation, finance, sports etc. In future, GAN could also be applied to various other kinds of e-commerce tasks such as product recommendation, targeting deals and simulation of future events.

\newpage

\bibliographystyle{ACM-Reference-Format}
\bibliography{eCommerce-body}

\end{document}